# Modelling Contractual Arguments

Chris Reed & Aspassia Daskalopulu

Department of Information Systems and Computing
Brunel University, UK.
E-mail: { Aspassia.Daskalopulu, Chris.Reed}@brunel.ac.uk

## 1. INTRODUCTION

One influential approach to assessing the "goodness" of arguments is offered by the Pragma-Dialectical school (p-d) (Eemeren & Grootendorst 1992). This can be compared with Rhetorical Structure Theory (RST) (Mann & Thompson 1988), an approach that originates in discourse analysis. In p-d terms an argument is good if it avoids committing a fallacy, whereas in RST terms an argument is good if it is coherent. RST has been criticised (Snoeck Henkemans 1997) for providing only a partially functional account of argument, and similar criticisms have been raised in the Natural Language Generation (NLG) community—particularly by Moore & Pollack (1992)—with regards to its account of intentionality in text in general.

Mann and Thompson themselves note that although RST can be successfully applied to a wide range of texts from diverse domains, it fails to characterise some types of text, most notably legal contracts. There is ongoing research in the Artificial Intelligence and Law community exploring the potential for providing electronic support to contract negotiators, focusing on long-term, complex engineering agreements (see for example Daskalopulu & Sergot 1997). The negotiation process, which is a lengthy cycle of proposal and counter-proposal, can be seen as inherently argumentative in nature with each party involved trying to influence the agreement in a way that best serves their own interests. The negotiation process is conducted by parties exchanging proposed drafts of the contract, where each draft represents an argument put forward by one party to persuade the other. Furthermore the internal structure of any given contractual document can be analysed as an implicit discussion where an implicit opponent makes requests for clarification and specification (particularly of contingencies that might arise). Supporting these aspects of contracts depends upon a rich model of the argumentative structure of the complex pre-contractual documents, and it is therefore disappointing that RST fails to account for such text.

It has also become clear (Reed 1998) that RST is fundamentally inappropriate for representing argument structure in three important respects: RST admits multiple analyses of a given piece of text and this is in direct contrast to the argumentation theoretic approach; particular structures that are frequently encountered in arguments are not catered for by RST; and finally, patterns of reasoning that underlie an argument (such as modus ponens, inductive generalisation and so on) can neither be represented by, nor inferred from an RST analysis (and even more so where multiple analyses exist).

This paper provides a brief introduction to RST and illustrates its shortcomings with respect to contractual text. An alternative approach for modelling argument structure is presented (extending Reed & Long 1997a) which not only caters for contractual text, but also overcomes the aforementioned limitations of RST. Finally it is shown that this approach meets the criticisms expressed by both Snoeck Henkemans (1997) and Moore and Pollack (1992) by offering a truly functional account of illocutionary purpose.

## 2. AN OVERVIEW OF RHETORICAL STRUCTURE THEORY

### 2.1 RST ASSUMPTIONS, METHODOLOGY AND BASIC CONCEPTS

Rhetorical Structure Theory (RST) developed by Mann and Thompson (1987; 1988) purports to evaluate text (including arguments) in terms of its coherence. The characteristics of RST as a descriptive framework for natural text are:

(i)  It describes relations between parts of text in functional terms, whether such relations are grammatically signalled or otherwise.
(ii)  It identifies hierarchical structure in text.
(iii)  Its scope is written monologue and it is insensitive to text size.

RST is put forward as a unifying framework, applicable to virtually any natural text of any size. An RST analysis of natural text operates within the following assumptions: The analyst has access to the particular text that is analysed, but no direct access to either the writer or the reader of such text. The analyst however knows the context in which the given text was written and shares the cultural conventions of both the reader and the writer of the text. The purpose of the analysis is to make explicit the function of the text along two dimensions, namely the writer's intention and the reader's comprehension; thus text is assessed on how effectively the writer's intentions are communicated to the reader.

The analysis is conducted by identifying relations between text spans (that is, uninterrupted linear intervals of text). A number of relations that can obtain between text spans have been identified by Mann and Thompson and are summarised in the following table:

| | |
|---|---|
| Circumstance | Antithesis and Concession |
| Solutionhood | Antithesis |
| Elaboration | Concession |
| Background | Condition and Otherwise |
| Enablement and Motivation | Condition |
|    Enablement | Otherwise |
|    Motivation | Interpretation and Evaluation |
| Evidence and Justify | Interpretation |
|    Evidence | Evaluation |
|    Justify | Restatement and Summary |
| Relations of Cause | Restatement |
|    Volitional Cause | Summary |
|    Non-Volitional Cause | Other Relations |
|    Volitional Result | Sequence |
|    Non-Volitional Result | Contrast |
|    Purpose | |

Table 1: Organization of the Relation Definitions (Mann & Thompson 1987)

Mann and Thompson note that the set of relations that they have identified is not necessarily complete and that additional relations may be added to that if the analyst finds that none of those serve his purpose adequately.

Each relation is defined between two non-overlapping text spans with one of these labelled the nucleus and the other as the satellite of the relation. Though RST does not provide an explicit direction about how these labels are decided it appears that the nucleus is the text span that contains essential information, in that its absence would reduce the meaningfulness of the text.

A relation definition comprises four fields: constraints on the nucleus (N), constraints on the satellite (S), constraints on the combination of nucleus and satellite (N+S) and the effect. For example the definition of the relation JUSTIFY is:

Relation Name: JUSTIFY
Constraints on N: none
Constraints on S: none
Constraints on N+S:
Reader's comprehending S increases Reader's readiness to accept Writer's right to present N.
The effect: Reader's readiness to accept Writer's right to present N is increased.
Locus of the effect: N.

To illustrate relation definitions further, consider another example, the definition of the relation ELABORATION:

Relation Name: ELABORATION
Constraints on N: none
Constraints on S: none
Constraints on N+S:
S presents additional detail about the situation or some element of subject matter which is presented in N or inferentially accessible in N in one or more of the ways listed below.
In the list if N presentes the first member of any pair, then S includes the second:
1.	set: member
2.	abstract: instance
3.	whole: part
4.	process: step
5.	object: attribute
6.	generalization: specific
The effect: Reader recognizes the situation presented in S as providing additional detail for N. Reader identifies the element of subject matter for which detail is provided.
Locus of the effect: N and S.

A relation between two text spans is pictorially represented by a structure diagram:

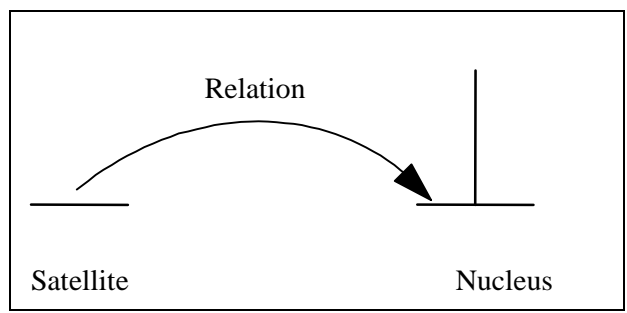

Figure 1: RST relation structure diagram

Each such relation is an elementary structure of the text that is analysed. Mutliple relations can be arranged into composite structures, that is patterns that define how a large span of text is analyzed in terms of other spans. Such composition of elementary relations is subject to the following constraints:

(i)     Completeness: The top level of the structure contains all the text spans constituting the entire text.
(ii)    Connectedness: Except for the entire text as a text span, each text span is either a minimal unit contributing as nucleus or satellite in a relation (elementary structure), or a constituent of a composite structure.
(iii)   Uniqueness: Each structure consists of a different set of text spans and each relation within a structure applies to a different set of text spans.
(iv)    Adjacency: The text spans of each structure constitute one text span.

As Mann and Thompson (1987) note completeness, connectedness and uniqueness taken in conjunction entail that RST analyses of texts yield hierarchical tree structures. The leaves of such a structure taken from left to right correspond to the entire text in the linear order in which they appear in it.

To illustrate these concepts RST analysis was conducted on a randomly chosen piece of text, in which text spans are numbered to facilitate reference:

1.  The wealth of societies in which the capitalist method of production prevails, takes the
2.  form of an "immense accumulation of commodities",
3.  wherein individual commodities are the elementary units.
4.  Our investigation must therefore begin with an analysis of the commodity.
5.  A commodity is primarily an external object,
6.  a thing whose qualities enable it, in one way or another, to satisfy human wants.
7.  The nature of these wants, whether for instance they arise in the stomach or the imagination, does not affect the matter.
8.  Nor are we here concerned with the question, how the thing satisfies human want, whether directly as a means of subsistence(that is to say, as an object of enjoyment), or indirectly as a means of production.

Example 1: Karl Marx, Capital, vol. 1, J. M. Dent & Sons Ltd.

The analysis of this text gave rise to the following hierarchical structure:

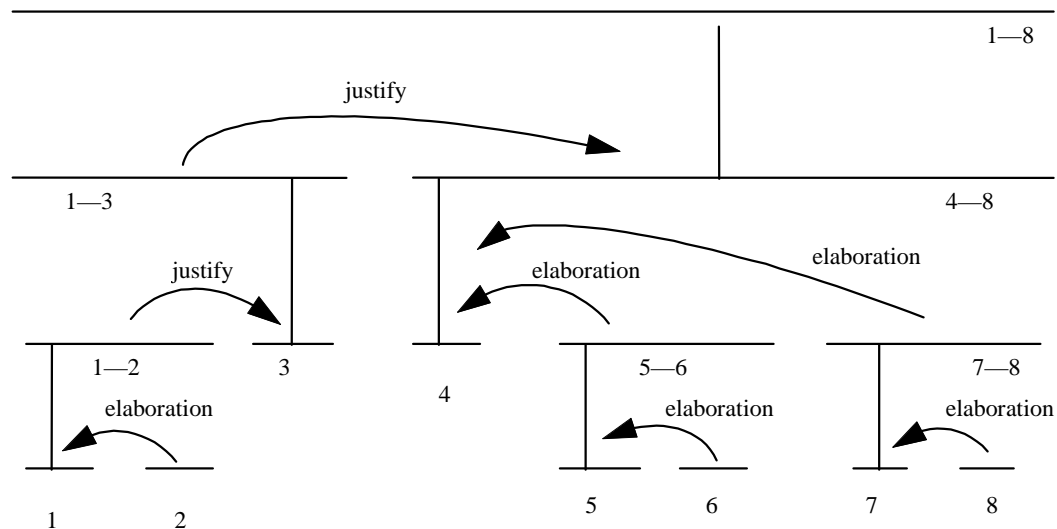

Figure 2: RST analysis of Marx example

2.2. OPERATIONALISATION OF RST

One powerful application of RST is to the field of natural language generation (NLG): if a system has a goal to communicate information to a hearer, how can that goal be fulfilled? RST offers a way of planning text by viewing each rhetorical relation as an operator—a step which has precisely defined requirements and effects. Text generation is then a task of creating a sequence of these operators such that the requirements of the first are true in the initial, pre-discourse state, and the effects of the last include the desired communicative goal (Hovy 1988). This sequence of rhetorical relations can then be refined to the appropriate grammatical and lexical form by more established realisation techniques.

## 3. CRITIQUE OF RST

### 3.1. FUNDAMENTAL PROBLEMS WITH RST

Although Rhetorical Structure Theory has been a highly popular technique in NLG (Hovy 1993), it has become clear from the demands of discourse generation that RST has a key failing with respect to the purported claims of functional adequacy. The conflation of informational (i.e. rhetorical, structural) and intentional (i.e. illocutionary) content leaves text generation systems without a means for recovering from communicative failure (such as the hearer misunderstanding) and answering follow-up questions (Moore & Pollack1992). More recently, this conflation has also been recognised as a problem for an RST-based analysis of argument: Snoeck Henkemans (1997) concludes that the account could at best be "partly functional". RST also suffers, however, from a more fundamental problem which becomes manifest in argument analysis. Despite Mann and Thompson's opening claim that "it is insensitive to text size", RST seems to be unable to adequately represent the high level abstract structure of argument. This intuitive shortcoming is a result of several assumptions upon which the account is founded. Mann and Thompson discuss the key role played by the notion of nuclearity - that relations hold between one nucleus and one satellite. They do, however, concede (YEAR?p269) that there are a few cases in which nuclearity breaks down - and these they regard as rather unusual. The two types of multi-nuclear constructs they identify are enveloping structures—"texts with conventional openings and closings"—and parallel structures— "texts in which parallelism is the dominant organizing pattern". Both of these are not just common in argument, but form key components. Enveloping structures are precisely what are described by, for example, Blair (1838), when presenting the dissection of argument into introduction, proposition, division, narration, argumentative, pathetic and conclusion (these are by no means obligatory in every argument, nor is there any great consensus over this particular characterisation; most authors, however, would agree that some such gross structure, usually involving introduction and conclusion, is appropriate). These structures are found with great frequency in natural argument, and cannot, therefore, be ignored. Parallel structures form the very basis of argument, since only the most trivial will involve lines of reasoning in which a single premise supports a single conclusion. Multiple subarguments conjoined to support a conclusion are the norm (see for example, (Cohen 1987), (Reed & Long 1997b) and these, necessarily form parallel structures.

Another point of dissonance between RST and argument analysis is that it is accepted that a text may be amenable to multiple RST analyses—not just as a result of ambiguity, but because there are, at a fundamental level, "multiple compatible analyses". Mann and Thompson (1987, p. 265) comment: "Multiplicity of RST analyses is normal, consistent with linguistic experience as a whole, and is one of the kinds of pattern by which the analyses are informative". This contrasts with the view in argumentation theory, where one argument has a single, unequivocable structure. There may, of course, be practical problems in identifying this structure, and two analysts may disagree on the most appropriate analysis (and indeed this latter has a close parallel in RST, since different analysts are at liberty to make different 'plausibility judgements' as to the aims of the speaker). The presence of these problems, however, is not equivalent to claiming that arguments may simply have more than one structure, a claim which would pose insurmountable problems to the evaluation process (the presence of inherent structural multiplicity would present the possibility of an argument being simultaneously evaluated as good and bad).

Finally, there is a more intuitive problem with RST, highlighted by analysing argument structure. Although there is much debate over the number and range of rhetorical relations (e.g. (Knott & Dale 1996), (Hovy 1993)) there seems to be no way of dealing with the idea of argumentative support. In the first place, as Snoeck-Henkemanns (1997) points out, Motivation, Evidence, Justification, Cause, Solutionhood and other relations could all be used argumentatively (as well, of course, as being applicable in non-argumentative situations). Thus it is impossible to identify an argumentative relation on the basis of RST alone. Secondly, RST offers no way of capturing higher level organisational units, such as Modus Ponens, Modus Tolens, and so on. For although their structure (or at least the structure of any one instance) can be represented in RST—and, given Marcu's (1996) elegant extensions, even their hierarchical use in larger units—adopting this approach necessitates a lower level view. It becomes no longer possible to represent and employ an MT subargument supporting the antecedent of an MP; rather, the situation can only be characterised as P supporting through one of the potentially argumentative RST relations Q, and showing that ~Q, so ~P, and ~P then supporting through one of the potentially argumentative RST relations R, therefore R. Apart from being obviously cumbersome, the representation has lost the abstract structure of the argument altogether, and is not generalisable and comparable to other similar argument structures. (It could perhaps be maintained that such structures could be represented as RST schemas, but there are several problems with such an approach: in the first place, schemas cannot abstract from individual relations, so there would need to be a separate 'MP' schema for each possible argumentative support relation; furthermore, the optionality and repetition rules of schema application (p248) are not suited to argument, as they license the creation of incoherent argument structure).

3.2. RST ANALYSIS OF CONTRACTUAL TEXT

Legislation and legal contracts have, in recent years, been the focus of much research mainly in the Artificial Intelligence community. A recent research project was concerned with the development of electronic tools to support contractual activity, especially negotiation of long-term, complex engineering agreements (Daskalopulu & Sergot 1997; Daskalopulu 1998). The negotiation of such contracts is a lengthy cycle of proposal and counter-proposal between two parties, and it can be seen as inherently argumentative in nature as each party tries to influence the agreement in a way that best serves their own interests. The negotiation is typically conducted by parties exchanging drafts of the proposed contract; each such draft may be regarded as an argument put forward by one party with the intention to persuade the other. Supporting such negotiation could benefit substantially by some means of assessing the communicative effect of contractual text. Moreover, establishing the functional roles of various contractual provisions within a contract is important for another aspect of contractual activity: in litigation situations the courts of law are supposed to rule for or against a party's motion by interpreting the agreement and trying to establish the parties' intentions at the time of making it, using contractual documents as a guide. Under the English law of contract (and to the best of our knowledge in most other legal traditions) the *parol evidence* rule applies, whereby in the presence of written contracts the text is taken to express all that the parties agreed and only that (Atiyah 1989). A court of law in a litigation situation is therefore concerned with establishing the writers' (the parties') intentions as these are manifested through the text they upon which they agreed.

Mann and Thompson (1987, p. 265) note: " Certain text types characteristically do not have RST analyses. These include laws, contracts, reports "for the record" and various kinds of language-as-art, including some poetry". The reasons for this inapplicability of RST to these kinds of text are not documented[1] by Mann and Thompson though.

In an effort to uncover such reasons a conventional RST analysis of contractual text is presented below. The experiment demonstrates not that RST is inapplicable to contractual text, but rather, that there are a number of important points. Figure 3 represents an RST analysis of an extract from an agreement on arbitration shown below.

1.1.   The arbitral tribunal shall be composed of three members,
1.2.   one to be appointed by each party
1.3.   and the third member, who shall act as president,
1.4.   to be appointed by the <appointing authority>.
2.1.   The member of the tribunal appointed by the first party shall be <name and address>
2.2.   The member appointed by the second party shall be <name and address>.
3.1.   If at any time a vacancy shall occur on the Tribunal
3.2.   by reason of the death, resignation, or incapacity for more than 60 days of any member, or for anyother reason,
3.3.   such vacancy shall be filled as soon as possible
3.4.   in the same manner as the original appointment of that position.

Example 2: Model Business Contracts, Croner Publications Ltd. 1988

---

[1] Although in the case of language-as-art or some poetry they might be obvious: it is not necessarily the case that the writer's intention is to convey some particular message to the reader, rather it might be to create a particular emotional effect with which the functional account of RST is not concerned.

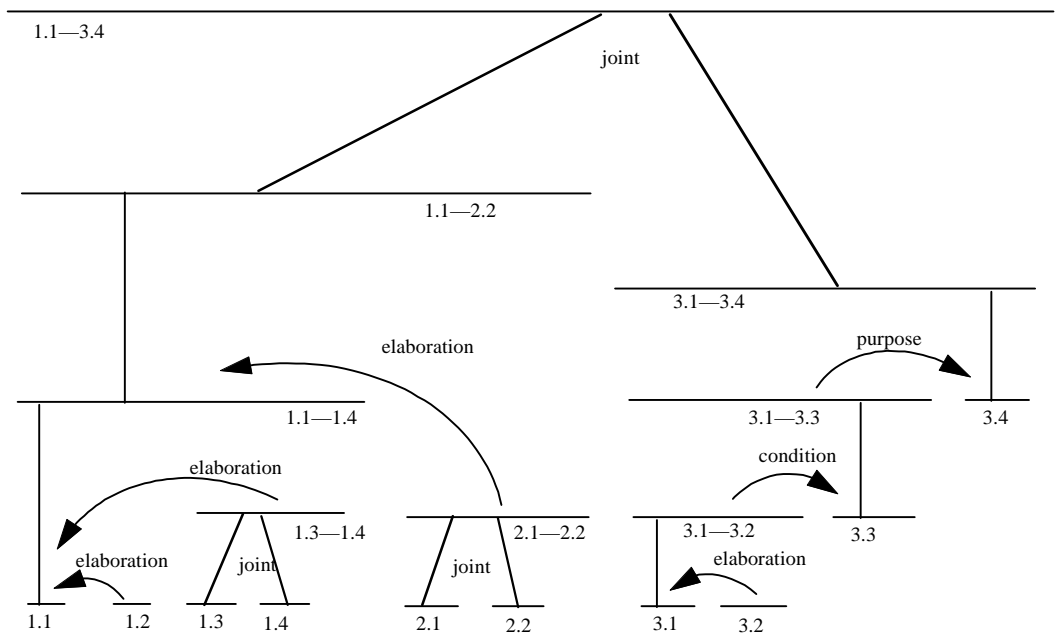

Figure 3: RST analysis of contract example

The RST analysis of example 2 highlights the central role of the analyst's judgement in identifying text spans and in determining which relation applies between them (incidentally, this was also the case for the Marx example). The text span comprising 1.3—1.4 could for example be regarded as providing BACKGROUND to 1.1. Similarly, 2.1—2.2 might have been regarded as being JOINT to 1.1—1.4[2]. Moreover the set of relations supported by RST is not necessarily complete; should none of the defined relations be deemed satisfactory to account for the relationship between two text spans, it seems that the analyst may make up a new one, as long as the definition conforms with the RST framework (by specifying all four of its fields). Mann and Thompson point out that the analyst has in effect to make plausibility judgements about the writer's intention and the reader's comprehension and this gives rise to multiple RST analyses for the same piece of text. In seeking a functional account of contractual text however negotiating parties and courts of law would require something more conclusive.

The functional account that is appropriate for contractual text (for the purposes mentioned earlier) is very different from the one provided by RST. The constraints for completeness, connectedness, uniqueness and adjacency imposed by RST result in tree-like structures for linear text with each text span having a unique effect (a unique functional role) within a single analysis. Contract documents are organized in a tree-like structure syntactically, that is they are organized in parts, where each part contains sections, and the latter contain provisions which can further be analyzed in terms of their constituent sentences and so on. Semantically however contract documents are organized as graphs, with a heavy amount of cross-referencing and provisions playing multiple roles. For example (cf. Daskalopulu & Sergot 1997) a contractual provision may be providing a definition for a term, prescribing duties and rights for the parties, specifying a procedure that needs to be followed for certain goals to be achieved (the contract example presented earlier contained such procedural specification) and so on. The functional account that is required for contractual text is therefore one that caters for non-linear text and allows one text span to participate in multiple relations reflecting the diverse functional roles it plays within the agreement.

---

[2] JOINT is actually a means of composing elementary structures into compound ones (a schema application in Mann and Thompson's terms). Here we treat it just as a vacuously defined relation, that is, there is no specification of constraints on nucleus, satellite or their combination and no effect. The result is identical to that of Mann and Thompson's.

Revisiting the contract example earlier, the following diagram illustrates the kind of functional account that is desirable:

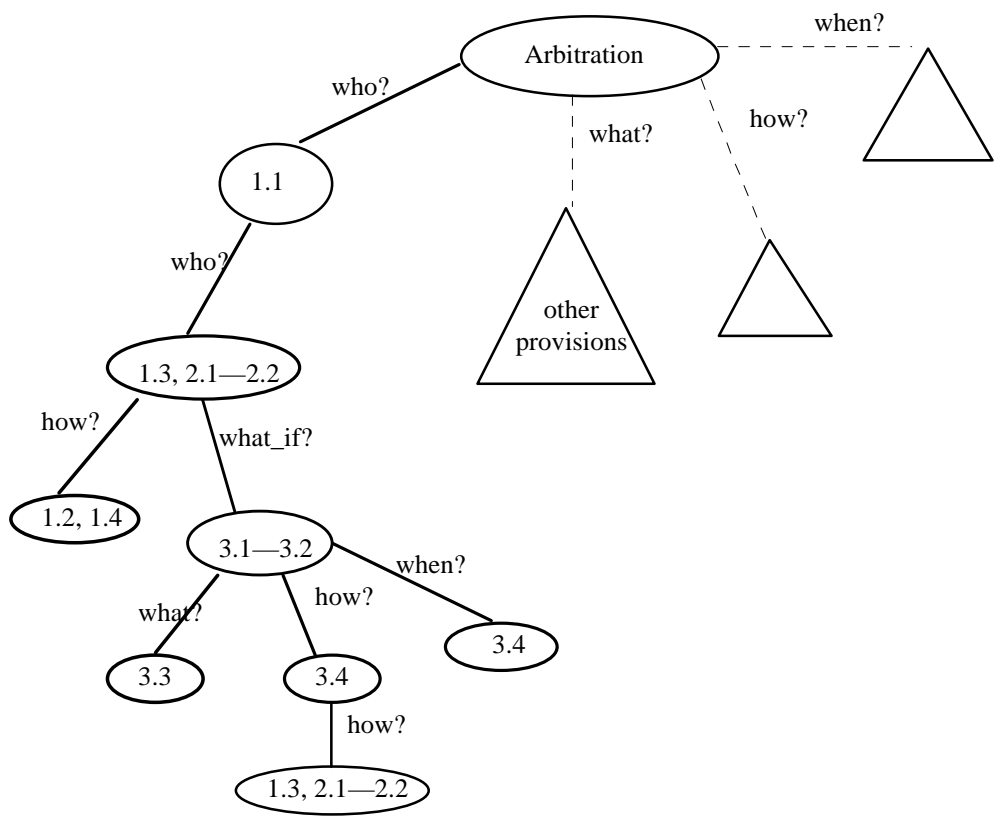

Figure 4: Functional analysis of contract example

The diagram shows the tree that corresponds to the graph for the text excerpt (that is, instead of repeating some nodes arcs essentially point to them directly). Each of the who/how/what/what_if arcs can be treated in a uniform way as a specification of various kinds. The functional account of a large agreement dealing with a multitude of issues (for example, billing and payment arrangements, force majeure provisions, quality monitoring and so on) is a collection of such interrelated structures.

Finally, though there is a persuasive nature to contracts—reflected in drafts exchanged by parties—with each participant trying in a competitive manner to secure the "best" terms for him, there is also a deliberative aspect: on a variety of issues the parties deliberate on the manner which is best suited to operationalise their agreement[3]. The contract example mentioned earlier is appropriate to illustrate this: parties are not in direct competition as regards the appointment of an arbitratory tribunal; rather they may argue for or against, say the number of members of the tribunal, or the time allowed for a vacancy to exist before it is filled, in an effort to cater for contingencies that might arise in the future. In effect they are arguing but not necessarily for their own narrow interests but rather for the best way that allows the business exchange to proceed smoothly. The approach proposed in the following section extends RST in a manner that enables both argumentative and deliberative accounts to be represented in a single framework.

4. A NEW APPROACH

To address the fundamental problems noted in section 2 and particularly the last one in section 2.1, and to provide a platform for representing the functional effects of contractual text, an alternative approach is proposed whereby RST is subsumed by a layer which explicitly represents argumentative constructs (Reed98), (Reed&Long97)(IJCAI). At this layer, support relations between propositions are reified, and are employed in defining the structure of argument. These structural relations are then operationalised to enable planning with operators encapsulating the various argument forms (MP, MT, inductive generalisation, etc.). The definitions of the operators make extensive use of intentional constructs thus avoiding the problems outlined by (Moore & Pollack 1992) (so that, e.g., the MP operator has the effect of increasing the hearer's belief in a proposition).

The argumentative structures represented at this abstract layer can be mapped on to the most appropriate set of RST relations (thus, for example, the implicature in an MP may be realised into any one of the potentially argumentative relations mentioned above). The approach thus maintains the generative capabilities of RST (particularly when extended along the lines of (Marcu 1996) to ensure coherency through adducement of canonical ordering constraints), whilst embracing the intuitive argumentative relationships at a more abstract level. It is these latter relationships which characterise the structure of the argument (i.e. the structure which argumentation theory strives to determine). The relationships are also unambiguous: a single argument has exactly one structure at this level abstraction (though multiplicity is not thereby prevented at the RST level). Further, parallelism occurs only at the higher level of abstraction (multiple subarguments contribute to a conclusion, but each subargument is mononucleaic), and similarly, enveloping structures are also characterised only at the higher level (thus the RST is restricted to a predominantly mononucleaic structure). Finally, complete argument texts are not obliged to have complete RST trees. For although most parts of a text are likely to have unifying RST analyses, and although there must be a single overarching structure at the highest level of abstraction, the refinement to RST need not enforce the introduction of rhetorical relations between parts. This expands the flexibility and generative capacity of the system encompassing a greater proportion of coherent arguments.

Though motivated by the requirements of sophisticated text generation, the model tackles many of the problems inherent to RST-only analysis. In particular, it offers a fully functional account by distinguishing the intentional and informational components of text structure, and answers Snoeck Henkemans criticisms by enabling argumentative relations between textual units to be handled explicitly. The structures generated by, and represented in, the system are essentially those characterised by Freeman (1991) as the 'standard treatment', whereby propositions can serve as premises or conclusions connected by convergent or linked support (it is recognised that there are, of course, much richer characterisations and diagrammatic techniques for investigating argument structures—Freeman himself develops one such—but the standard treatment offers a simple, tractable, and sufficiently expressive account to be of great interest).

---

[3] This distinction between notions of persuasion and deliberation is adopted from (Walton and Krabbe 1995).

Although the work in (Reed98), (Reed&Long97) focuses specifically upon persuasive argument, the same approach can be adopted towards the inherently deliberative internal structure of parts of a contract. In particular, that structure can be represented diagramatically using nodes to represent propositions and arcs to represents relations between them. In the same way that a persuasive argument can be seen as an implicit dialogue, whereby each statement of the writer has been elicited by some implicit question (of relevance or ground adequacy), a contract too can be viewed as inherently dialectical, whereby an implicit opponent may offer questions forcing specification: the who question demanding role instantiation; the when question demanding temporal specification; the how question demanding specification of means; and, most frequently, the what-if question, demanding specification of contingency action. It is these questions which characterise the relationships between nodes in the contract graph. With an isomorphic relationship between the structure of persuasive discourse and that of deliberative discourse, the techniques developed for computational representation of the former can also be applied to the latter.

## 5. CONCLUSIONS AND FUTURE WORK

Rhetorical Structure Theory, though a competent model of small scale text structure with wide applicability in both discourse analysis and natural language generation, suffers from a range of problems many of which become insurmountable when considering its application to large scale arguments and contracts. A more abstract level of representation, subsuming RST, is required to provide a functional account of the complex structure and interdependencies present in such text. The representation developed for handling the structure of persuasive text has been shown to cope with contractual text as a result of an isomorphism in the structure of the two genres, and in particular, that it can be appropriate to view each as an implicit discussion.

Current work is exploring in more detail the practical advantages such a computational representation may afford. In particular, a means of representing and manipulating the large scale structure of a contract may be of use in supporting the drafting, negotiation and litigation activities through provision of a tool for navigation and referencing of a large contractual agreement (such agreements may often run to hundreds of pages and have a dynamic nature running over many years). An integration with the work of (Daskalopulu & Sergot 1995), and with others working on legal information systems thus represents a potentially fruitful avenue of investigation. A more ambitious aim is to extend the model presented in (Reed 1998) to cover the automatic generation of contract structure, fulfilling either a role of critic of human generated contracts, or one of preliminary authoring in well defined domains.

| | |
|---|---|
| Circumstance | Antithesis and Concession |
| Solutionhood | Antithesis |
| Elaboration | Concession |
| Background | Condition and Otherwise |
| Enablement and Motivation | Condition |
|     Enablement | Otherwise |
|     Motivation | Interpretation and Evaluation |
| Evidence and Justify | Interpretation |
|     Evidence | Evaluation |
|     Justify | Restatement and Summary |
| Relations of Cause | Restatement |
|     Volitional Cause | Summary |
|     Non-Volitional Cause | Other Relations |
|     Volitional Result | Sequence |
|     Non-Volitional Result | Contrast |
|     Purpose | |

Table 2: Organization of the Relation Definitions (Mann & Thompson 1987)

Reed C. A. & Daskalopulu A. (1998). Modelling Contractual Arguments. Proceedings of the 4th International Conference on Argumentation (ISSA-98).SICSAT, pp. 686–692.

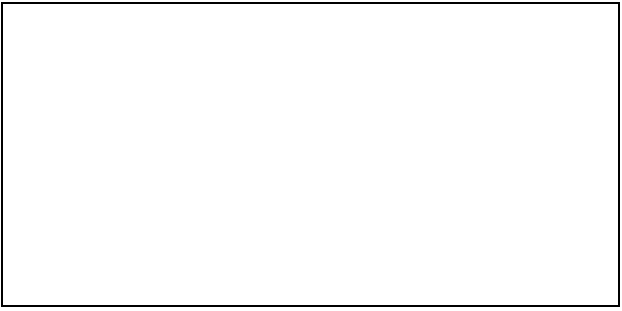

Figure 5: RST relation structure diagram

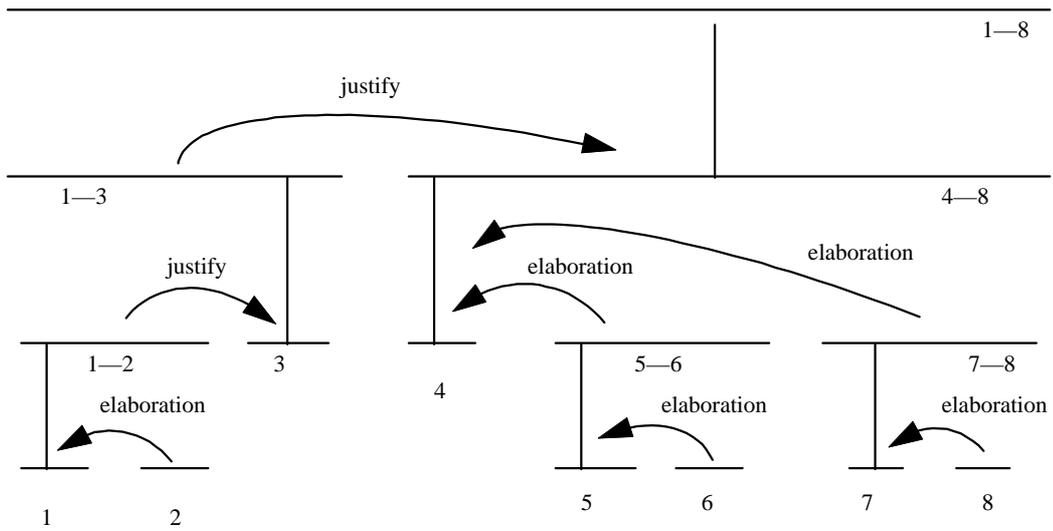

Figure 6: RST analysis of Marx example

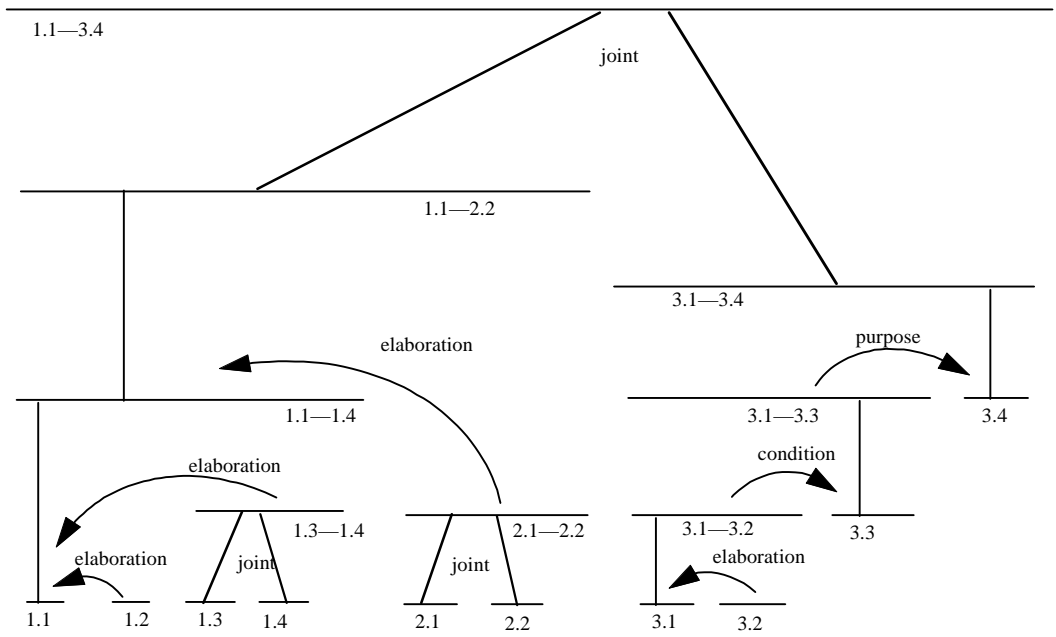

Figure 7: RST analysis of contract example

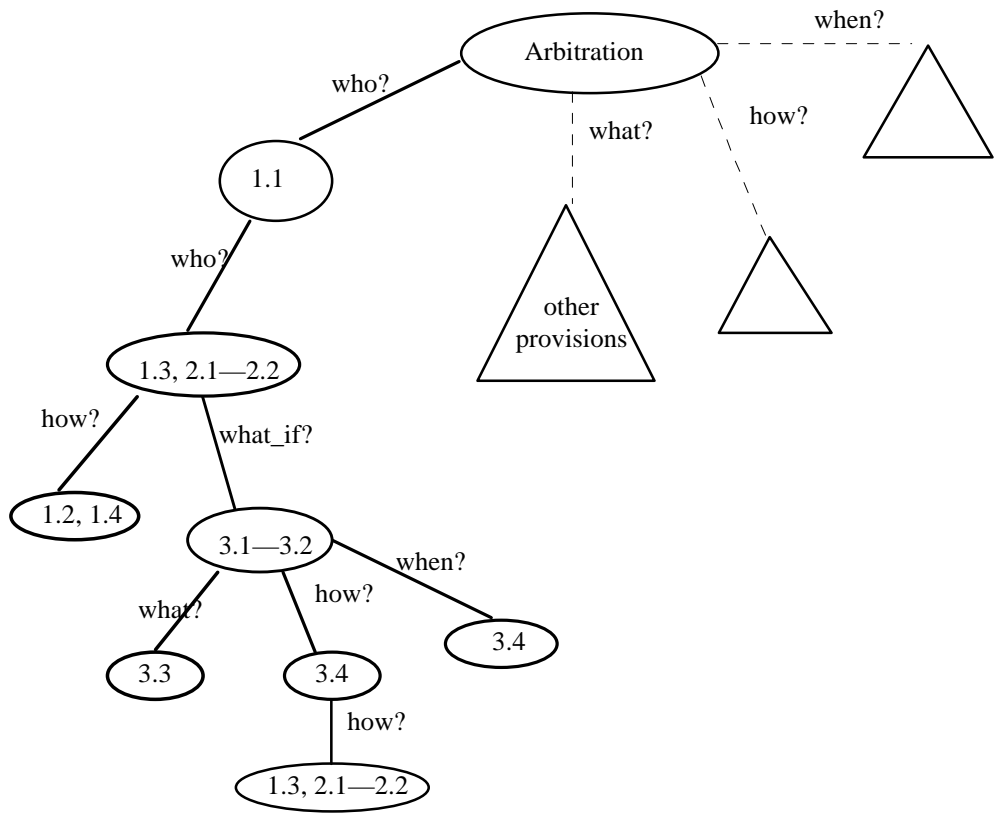

Figure 8: Functional analysis of contract example